\def\year{2022}\relax
\documentclass[letterpaper]{article} 
\usepackage{aaai22}  
\usepackage{times}  
\usepackage{helvet}  
\usepackage{courier}  
\usepackage[hyphens]{url}  
\usepackage{graphicx} 
\usepackage{natbib}  
\usepackage{caption} 
\DeclareCaptionStyle{ruled}{labelfont=normalfont,labelsep=colon,strut=off} 
\usepackage{algorithm}
\usepackage{algorithmic}

\usepackage{newfloat}
\usepackage{listings}
\floatstyle{ruled}
\newfloat{listing}{tb}{lst}{}
\floatname{listing}{Listing}
\title{Hidden Complexities in the Computational Modeling of Proportionality for Robotic Norm Violation Response}
\author{
    Ruchen Wen and Tom Williams 
}
\affiliations{
    MIRRORLab \\
    Colorado School of Mines\\

    Golden, CO 80401\\
    \{rwen,twilliams\}@mines.edu
}

\copyrighttext{Presented at the AI-HRI Symposium at AAAI Fall Symposium Series (FSS) 2022}

\begin{document}

\maketitle

\begin{abstract}
Language-capable robots hold unique persuasive power over humans, and thus can help regulate people's behavior and preserve a better moral ecosystem, by rejecting unethical commands and calling out norm violations. However, miscalibrated norm violation responses (when the harshness of a response does not match the actual norm violation severity) may not only decrease the effectiveness of human-robot communication, but may also damage the rapport between humans and robots. Therefore, when robots respond to norm violations, it is crucial that they consider both the moral value of their response (by considering how much positive moral influence their response could exert) and the social value (by considering how much face threat might be imposed by their utterance). In this paper, we present a simple (naive) mathematical model of proportionality which could explain how moral and social considerations should be balanced in multi-agent norm violation response generation. But even more importantly, we use this model to start a discussion about the hidden complexity of modeling proportionality, and use this discussion to identify key research directions that must be explored in order to develop socially and morally competent language-capable robots. 
\end{abstract}

\section{Introduction and Motivation}
Research has shown that language-capable robots hold unique persuasive power over humans. Robots are not only capable of influencing humans to comply with their requests and commands~\cite{bartneck2010influence,cormier2013would,rea2017wizard}, but can also exert moral influence over human moral norm systems~\cite{jackson2019language,jackson2018robot,williams2018bayesian}.
It is thus important for robots to use their persuasive power in a way that is going to help regulate people's behavior and preserve a better moral ecosystem, by rejecting unethical commands~\cite{briggs2021and,jackson2019language,wen2022teacher} and calling out norm-violating behavior~\cite{winkle2022norm,winkle2021boosting,jung2015using}.
For example, imagine a scenario where a robot and a team of humans are working together on a collaborative task, and one of the human teammate verbally abuses another team member who accidentally made a mistake. To exert positive moral influence, the robot might be expected to call out the norm violation (i.e., the insult) in order to helpfully mediate the team's dynamics.

Moreover, research has shown that robots are more persuasive when they act in a socially interactive~\cite{fong2003survey}, sociable~\cite{breazeal2004designing} or socially agentic manner~\cite{jackson2021theory}, by leveraging human-like social cues~\cite{chidambaram2012designing,ghazali2019assessing} and strategies~\cite{srinivasan2016help}. This is critical when considering how robots should be leveraging their persuasive power in moral context. When rejecting immoral commands, for example, if robots only focus on explaining the morality of actions without considering the social strategies they could be using during that communication, their communication could be less effective, and they themselves could be viewed less favorably. For instance, Jackson et al.~\cite{jackson2019tact} found that if robots give miscalibrated norm violation responses (i.e., the harshness of the responses does not match the actual norm violation severity), they will be perceived less favorably.

To appropriately calibrate the harshness of their norm violation responses, robots can use the same types of strategies that humans use, including social norms like \textit{Politeness}. In Brown and Levinson's Politeness Theory~\cite{brown1987politeness}, humans regularly negotiate the level of threat to one another's \textit{Face}, which is the public image that a person wants to preserve and enhance in front of others~\cite{brown1987politeness}. Face includes Positive Face, which is a person's wish for a desirable self-image, and Negative Face, which is a person's wish to be free from imposition and to have freedom of action~\cite{brown1987politeness}. 

From the Politeness Theory perspective, calling out a norm violation can be highly face-threatening, as it threatens both Positive Face (by making the violator looks bad in front of others) and Negative face (by appearing to intend to control the violator's behavior). 
Thus, it is crucial for robots to address norm violations in a way that the moral value (i.e., how much positive moral influence their response could exert) is proportional to the social value (i.e., how much face threat might be imposed by their utterance), in order to exert positive moral influence and avoid exerting unintentional negative moral influence.

Indeed, proportionality is one of the fundamental and universal moral motives underlying human social-relational psychology~\cite{rai2011moral}, which is ``the motive for rewards and punishments to be proportionate to merit, benefits to be calibrated to contributions, and judgments to be based on a utilitarian calculus of costs and benefits''. 
To the best of our knowledge, however, there has been no previous work on computational modeling of how humans reason about proportionality during communication -- or how robots could do the same. 

In this paper, we begin to consider how such a computational model might be developed, and show the substantial complexity and nuance behind what might at first glance seem a simple problem. We present a simple (naive) mathematical model of proportionality which could explain how moral and social considerations should be balanced in multi-agent norm violation response generation. We then analyze the components of this model and procedurally explain how each is more complex than might immediately meet the eye. Through this process, we identify key research directions that must be explored in order to develop socially and morally competent language-capable robots.

\section{A Preliminary Model of Proportionality}
To begin, let us consider how we might model the utility of a candidate speech act that could be used to address a norm violation in a multi-agent context. 
Given our desire for proportionality, we can start by assuming that overall utility should include components for both the moral benefits of the response, and the social benefits (that is, the negation of the social harms). This model can be represented as:


$$\mathcal{U_{A}} = \mathcal{U_{MA}} + \mathcal{U_{SA}}$$

where $\mathcal{U_{A}}$ denotes the utility of a speech act $\mathcal{A}$, $\mathcal{U_{MA}}$ denotes the \textit{moral} utility of a speech act $\mathcal{A}$, and $\mathcal{U_{SA}}$ denotes the \textit{social} utility of a speech act $\mathcal{A}$. 

\textit{Moral utility} is positively related to moral benefits. In potentially multi-agent contexts in which robots need to address norm violations in front of zero or more human observers, the moral benefits of the robots' response come from accurately correcting misconceptions for each observer (including the violator). 
Thus, the calculation for moral utility can be represented as:

$$\mathcal{U_{MA}} = \sum\limits_{i=1}^{\mid O\mid} (|S_a - S_{i}| - |S_a - S_c|)$$

Here, $\mid O \mid$ denotes the total number of observers (including the norm violator), $S_a$ denotes the actual norm violation severity.
This could be learned from human data~\cite{sarathy2017learning,wen2020dempster}, learned in one shot through language, or calculated through some type of utilitarian analysis. 
$S_c$ denotes the norm violation severity that the speech act $\mathcal{A}$ is trying to convey. This could be estimated from empirical evidence of human perceptions or other computational models of politeness theory~\cite{danescu2013computational}). 
Finally,  $S_i$ denotes the norm violation severity that is perceived by observer $O_i$ (the $i$-th observer).
Each observer's prior beliefs about a norm violation's severity could be estimated from prior behaviours from that specific observer. 

We might also expand this to: 

$$\mathcal{U_{MA}} = \sum\limits_{i=1}^{\mid O\mid} ((|S_a - S_{i}| - |S_a - S_c|) - \beta|S_a - S_c|).$$

Here, $\beta$ denotes the weight of the dishonesty penalty, which might be affected by the robot's role. For example, dishonesty might be more greatly penalized for a teacher robot than a tour guide robot. 
To maximize moral utility, robots thus need to be honest with the norm violation severity: the norm violation severity they are trying to convey should be exactly (or very close to) the actual norm violation severity.

\textit{Social utility} is negatively related to the amount of face threat. The more people there are to observe a face threat (and the more the violator cares about their perception by each observer) the great the social penalty they will suffer from a harsh norm violation response.
Thus, the calculation for social utility can be represented as:
$$\mathcal{U_{SA}} = - \sum\limits_{i=1}^{\mid O \mid} (I (O_i,v) \times F(v,\mathcal{A})).$$

Here, $I (O_i,v)$ denotes the importance for a norm violator $v$ to maintain a positive social image in front of the observer $O_i$.
This could be estimated from the relational/organizational hierarchy or perceived power dynamic between the violator and the observer.
$F(v,\mathcal{A})$ denotes the amount of face threat the speech act $\mathcal{A}$ imposes to the violator $v$. This could be estimated from empirical evidence of human perceptions of language of different types, or through computational models of politeness theory~\cite{danescu2013computational}).


\section{Additional Sources of Complexity}
The simple model presented in the preceding section aims to balance moral benefits against the loss of social benefits. 
While this model seems, on its face, to capture the basics of proportionality, it obscures a wealth of complex considerations that would need to be addressed in practice. In this section, we will discuss these hidden complexities, and speculate as to how these considerations might need to be captured in a more complex model.

\subsection{Different types of observers}
In the proposed model, we treat every observer equally, in terms of contributing to the total moral and social utilities. In real-life scenarios, there may be different types of observers, who may need to be treated differently from other observers. In addition to bystanders as naively considered in the preliminary model, we should also consider other types of special observers:
\begin{itemize}
    \item The norm violator; 
    \item People who are (or will be) negatively impacted by the norm violation, (i.e., victims); and
     \item People who might not be aware of the norm and have facilitated or conducted the same norm violation.
\end{itemize}

By considering the different types of observers, we can understand the ways our model might need to be adjusted accordingly. 
For example, it may be more important to focus on correcting the violator than to worry about correcting (likely unobserved) misconceptions that could be held by other observers. If this is the case, we might need to assign different weights on the benefits of correcting different observers.

We also may need to consider face threat to observers beyond the violator. For example, when a norm violation is called out, some observers may also feel face threaten or experience other kinds of social awkwardness if they are not aware of the norm. In this case, the total amount of face threat a speech act may cause should also include the potential face threat that the observers may receive. 

Finally, victims are also important to consider when we calculate the utility of a norm violation response. For instance, calling out a norm violation, in general, may help the victims to reduce (or even avoid) the potential harm that the norm violation may cause to them. However, victims may not want somebody else to speak for them in certain cases, which might hurt both their positive and negative face by making them look incompetent and taking away their chances to defend themselves. Therefore, besides the correction benefits, the model should take all of these dimensions into account for calculating moral and social utilities.







\subsection{Scaling of face threat with the number of observers}
In the proposed model, the total amount of face threat is simply the summation of values over all observers. However, in reality, the amount of face threat does not tend to increase linearly with the number of observers. 
For example, losing face in front of 102 people and 112 people might be different, but the size of this difference is likely not as large  as the difference between losing face in front of two people and losing face in front of twelve people.
Given the fact that face threat likely scales nonlinearly with the number of observers, models of proportionality may need to include discount factors into their utility models.

\subsection{Potential benefits of face threat}
When calculating the social utility of a speech act, we consider all face threats to the violator as negative impacts and try to minimize face threat to increase utility. While this may seem reasonable at first glance, it may not always be truly beneficial to minimize face threat to a violator. First, if the violator has already caused harm, some face threat to the violator might be beneficial to everybody emotionally, for seeing the violator being called out (especially in the presence of the victims). Also, from the Confucian Ethical perspective, receiving face-threatening responses (e.g., blame-laden moral rebukes) for norm violations could help violators to cultivate their ``heart of shame''~\cite{zhu2020blame}, which is one of the key components of Confucian moral self-cultivation. Moreover, from a pedagogical perspective, harsh responses might create a stronger impression and thus be more effective in helping people learn and grow.

While a certain amount of face threat to the norm violator could potentially be beneficial, adding it to the moral utility model is still challenging. It is unclear how much moral benefit this face threat provides; and indeed, over-weighting this benefit could have clear negative effects (e.g., robots that intentionally seek opportunities for public shaming). Models of proportionality may require careful calibration of ostensible benefits of violator-directed face threats, and how this calibration may depend on various relational and social contextual factors.



\section{Discussion}

So far we have discussed a list of possible source of complexity in the modeling of proportionality. In fact, those complexities are not only important for generating norm violation responses, but also for other type of moral communication and other aspects of moral competence, such as moral reasoning and decision making. Given the relevance to those broader topics, it is thus worth discussing the more general concerns and challenges of enabling moral competence in social robots. 

In this study, we assumed that robots already had a substantial amount of prior knowledge about moral and social norms (i.e., some ground truth). However, where these norms come from and how robots should acquire this prior knowledge is an open question. 
Because people who have different culture backgrounds often have adhere to different moral and social norms (or do so in different ways), 
one potential solution is to use participatory design to collect information about which specific norms \textit{particular communities} would follow in different situations.  
Such an approach would be compatible with recent calls to explore Design Justice~\cite{costanza2020design,ostrowski2022ethics} and Engineering Justice~\cite{leydens2017engineering,williams2021human} approaches to robot design.
Even people from the same culture may have different moral and social beliefs due to individual life experience. In some cases, these differences might cause misunderstandings and conflicts, while in other cases, they might not negatively affect interpersonal communication. Thus, it is a challenge that robots should not only understand these differences, but more importantly, know when to seek common ground and when to preserve differences.

Additionally, in this work, our intention is to consider how we might enable robot competence to exert positive moral influence in order to help preserve and cultivate more harmonious human-robot eco-systems. While having robots calling out norm violations could be an effective approach to achieve our goals, it also bring concerns about robots ``norm policing''. A long term goal of this research area must be to balance the real need to call out norm violation responses without further turning robots into surveillance machines or developing tools that could be misused by law enforcement and other forms of state oppression.


\section{Summary}
In this paper, we presented a simple (naive) mathematical model of proportionality which could explain how moral and social considerations should be balanced in multi-agent norm violation response generation. We then used this model as a starting point to consider the hidden complexity of modeling proportionality. These considerations discussed above represent key research directions that must be explored in order to develop socially and morally competent language-capable robots. 

%



\section{Acknowledgments}
This work was funded in part by Young Investigator award FA9550-20-1-0089 from the United States Air Force Office of
Scientific Research.

\bibliography{ref}

\end{document}